\documentclass[10pt,twocolumn,letterpaper]{article}

\usepackage{wacv}
\usepackage{times}
\usepackage{epsfig}
\usepackage{graphicx}
\usepackage{amsmath,bm}
\usepackage{amssymb}
\usepackage{makecell}
\usepackage{tabularx}
\usepackage{booktabs}
\usepackage{multirow}

\usepackage{mathtools, stmaryrd}
\usepackage{xparse} \DeclarePairedDelimiterX{\Iintv}[1]{\llbracket}{\rrbracket}{\iintvargs{#1}}
\usepackage[accsupp]{axessibility}  % Improves PDF readability for those with disabilities.

\newcommand{\V}[1]{{\bm{#1}}}
\newcommand{\Vh}[1]{{\hat{\bm{#1}}}}

\def\wacvPaperID{276} % Enter the WACV Paper ID here

\wacvalgorithmstrack   % Uncomment this line if you are submitting to the Algorithms Track.
\wacvfinalcopy % *** Uncomment this line for the final submission

\ifwacvfinal
\usepackage[breaklinks=true,bookmarks=false]{hyperref}
\else
\usepackage[pagebackref=true,breaklinks=true,colorlinks,bookmarks=false]{hyperref}
\fi

\usepackage[capitalize]{cleveref}
\crefname{section}{Sec.}{Secs.}
\Crefname{section}{Section}{Sections}
\Crefname{table}{Table}{Tables}
\crefname{table}{Tab.}{Tabs.}

\begin{document}

%%%%%%%%% TITLE
\title{Full Contextual Attention for Multi-resolution  Transformers in Semantic Segmentation} % Replace with your title

\author{Loic Themyr$^{1,2}$ \and Clement Rambour$^{1}$ \and Nicolas Thome$^{1,3}$ \and Toby Collins$^{2}$ \and Alexandre Hostettler$^{2}$ \\
$^{1}$Conservatoire National des Arts et Métiers, Paris, France \\
$^{2}$IRCAD, Strasbourg, France \\
$^{3}$Sorbonne Université, CNRS, ISIR, F-75005 Paris, France \\
{\tt\small loic.themyr@lecnam.net}}

%Institution1\\
%Institution1 address\\
%{\tt\small firstauthor@i1.org}
% For a paper whose authors are all at the same institution,
% omit the following lines up until the closing ``}''.
% Additional authors and addresses can be added with ``\and'',
% just like the second author.
% To save space, use either the email address or home page, not both
%\and
%Second Author\\
%Institution2\\
%First line of institution2 address\\
%{\tt\small secondauthor@i2.org}
%}

%\author{First Author\\
%Institution1\\
%Institution1 address\\
%{\tt\small firstauthor@i1.org}
% For a paper whose authors are all at the same institution,
% omit the following lines up until the closing ``}''.
% Additional authors and addresses can be added with ``\and'',
% just like the second author.
% To save space, use either the email address or home page, not both
%\and
%Second Author\\
%Institution2\\
%First line of institution2 address\\
%{\tt\small secondauthor@i2.org}
%}

\maketitle
\thispagestyle{empty}

%%%%%%%%% ABSTRACT
\begin{abstract}
Transformers have proved to be very effective for visual recognition tasks. In particular, vision transformers construct compressed global representations through self-attention and learnable class tokens. Multi-resolution transformers have shown recent successes in semantic segmentation but can only capture local interactions in high-resolution feature maps. This paper extends the notion of global tokens to build GLobal Attention Multi-resolution (GLAM) transformers. GLAM is a generic module that can be integrated into most existing transformer backbones. GLAM includes learnable global tokens, which unlike previous methods can model interactions between all image regions, and extracts powerful representations during training. Extensive experiments show that GLAM-Swin or GLAM-Swin-UNet exhibit substantially better performances than their vanilla counterparts on ADE20K and Cityscapes. Moreover, GLAM can be used to segment large 3D medical images, and GLAM-nnFormer achieves new state-of-the-art performance on the BCV dataset.
\end{abstract}

%%%%%%%%% BODY TEXT
\section{Introduction}

Transformers have achieved state-of-the-art performances in various Natural Language Processing (NLP) tasks~\cite{AIAYN}. Recently, fully transformer-based models have reached excellent performances on vision tasks such as image classification~\cite{dosovitskiy2020vit} and semantic segmentation~\cite{SETR}.

The main appeal of transformers is their ability to grasp long-range interactions, which is a crucial point for semantic segmentation. However, this strategy is not easily scalable to high-resolution images involving a large number of patches, due to the quadratic complexity of the transformer's attention module. A simple and efficient strategy to tackle this limitation is to rely on multi-resolution approaches, where the attention in high-resolution feature maps is computed on sub-windows. There have been various recent attempts in this direction~\cite{liu2021Swin,wang2021pyramid,zhang2021aggregating,wang2021pvtv2,cao2021swinunet}. However, they limit the interactions of high-resolution features to within each window.

\begin{figure*}
    \centering
    \includegraphics[width=0.8\linewidth]{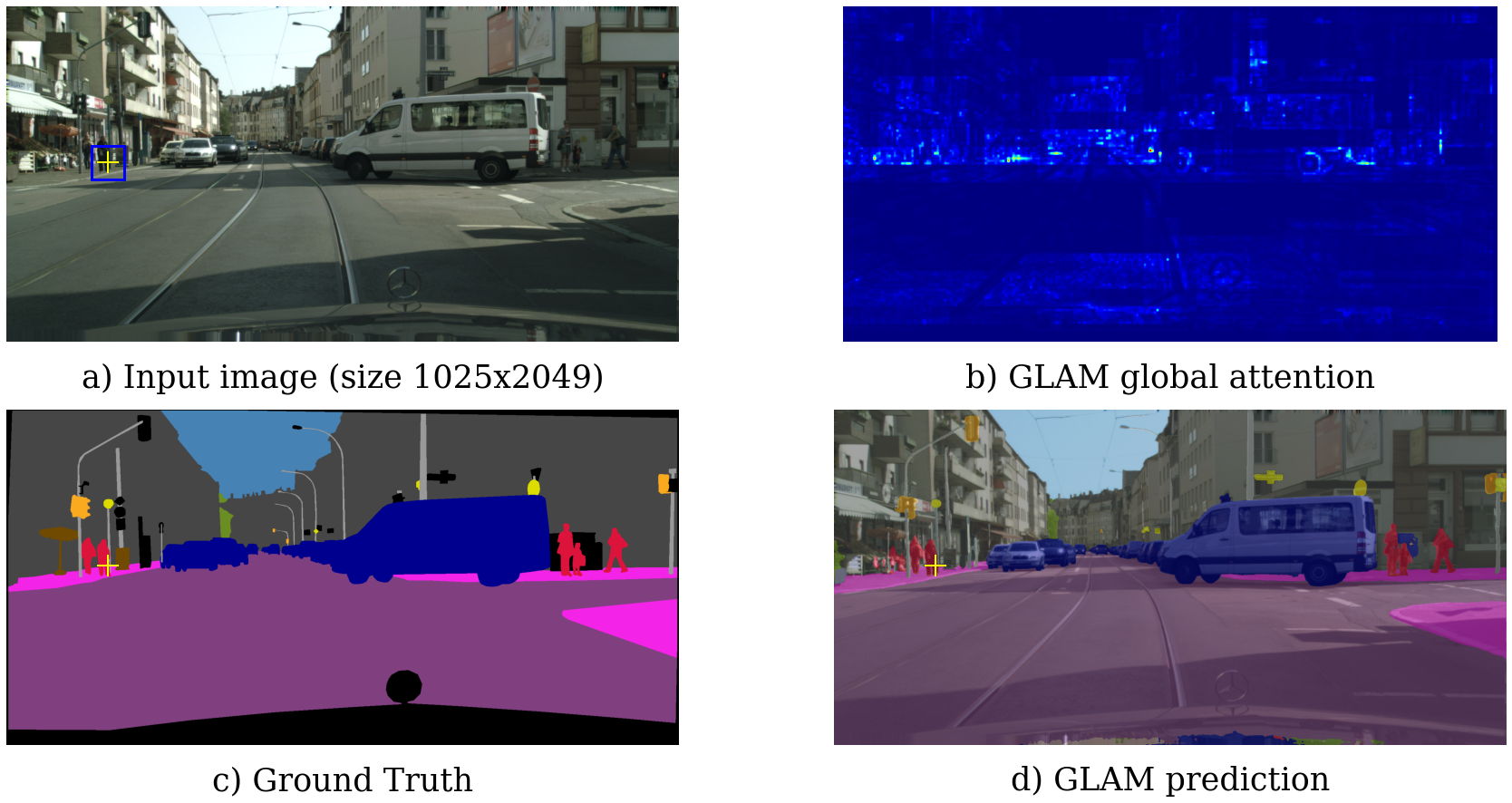}
    \caption{When segmenting the high-resolution image in a) with state-of-the-art multi-resolution transformers, \eg Swin~\cite{liu2021Swin}, the attention in the highest-resolution feature maps is limited to a small spatial region, \ie the blue square for the yellow-crossed pedestrian. Our method incorporates GLobal Attention in Multi-resolution transformers (GLAM). The GLAM attention map for the pedestrian in a) is depicted in b): it captures both fine-grained spatial information and long-range interactions, enabling successful segmentation, as shown in d).}
    \label{fig:intro}
\end{figure*}

We introduce an approach for semantic segmentation that incorporates global attention in multi-resolution transformers (GLAM). The GLAM module enables full-range interactions to be modeled at all scales of a multi-resolution transformer. As illustrated in Fig.~\ref{fig:intro}, incorporating GLAM into the Swin architecture~\cite{liu2021Swin} enables to jointly capture fine-grained spatial information in high-resolution feature maps and global context, where both elements are crucial for proper segmentation in complex scenes. This concept is illustrated in Fig.~\ref{fig:intro} where Fig.~\ref{fig:intro}a) shows an input image, and Fig.~\ref{fig:intro}b) shows the self-attention map provided by GLAM in the highest-resolution feature map for the pedestrian region pointed out by the yellow cross in Fig.~\ref{fig:intro}a). We can see that the attention map involves long-range interactions between other visual structures (cars, buildings), in contrast to the Swin baseline, where the window attention at a high-resolution feature map is limited to the small rectangular region in Fig.~\ref{fig:intro}a).
Consequently, GLAM has exploited longer-range interactions to successfully segment the image, as shown in~\ref{fig:intro}d).

%\noindent 
To achieve this goal, we have made the following novel contributions: 

\begin{itemize}
    \item We introduce the GLAM transformer, able to represent full-range interactions between all local features at all resolution levels. The GLAM transformer is based on learnable global tokens interacting between all visual features. To fully take into account the global context, we also design a non-local upsampling scheme (NLU).
    \item GLAM is a generic module that can be incorporated into any multi-resolution transformer. It consists of a succession of two transformers applied on the merged sequence of global and visual tokens and in-between global tokens. We highlight that the GLAM transformer can represent full-range interactions between image regions at all scales while retaining memory and computational efficiency. Beyond spatial interactions, global tokens also model the expected scene composition.
    \item Experiments on various generic (ADE20K), autonomous driving (Cityscape) and medical (Synapse) datasets show the important and systematic gain brought by GLAM when included into existing state-of-the-art multi-resolution transformers including Swin, Swin-Unet, and nn-Former. We also show that GLAM outperforms state-of-the-art methods on Synapse. Finally, ablation studies, model analysis, and visualizations are presented to assess the behavior of GLAM.   
\end{itemize}

\begin{figure*}
    \centering
    \includegraphics[width=.84\linewidth]{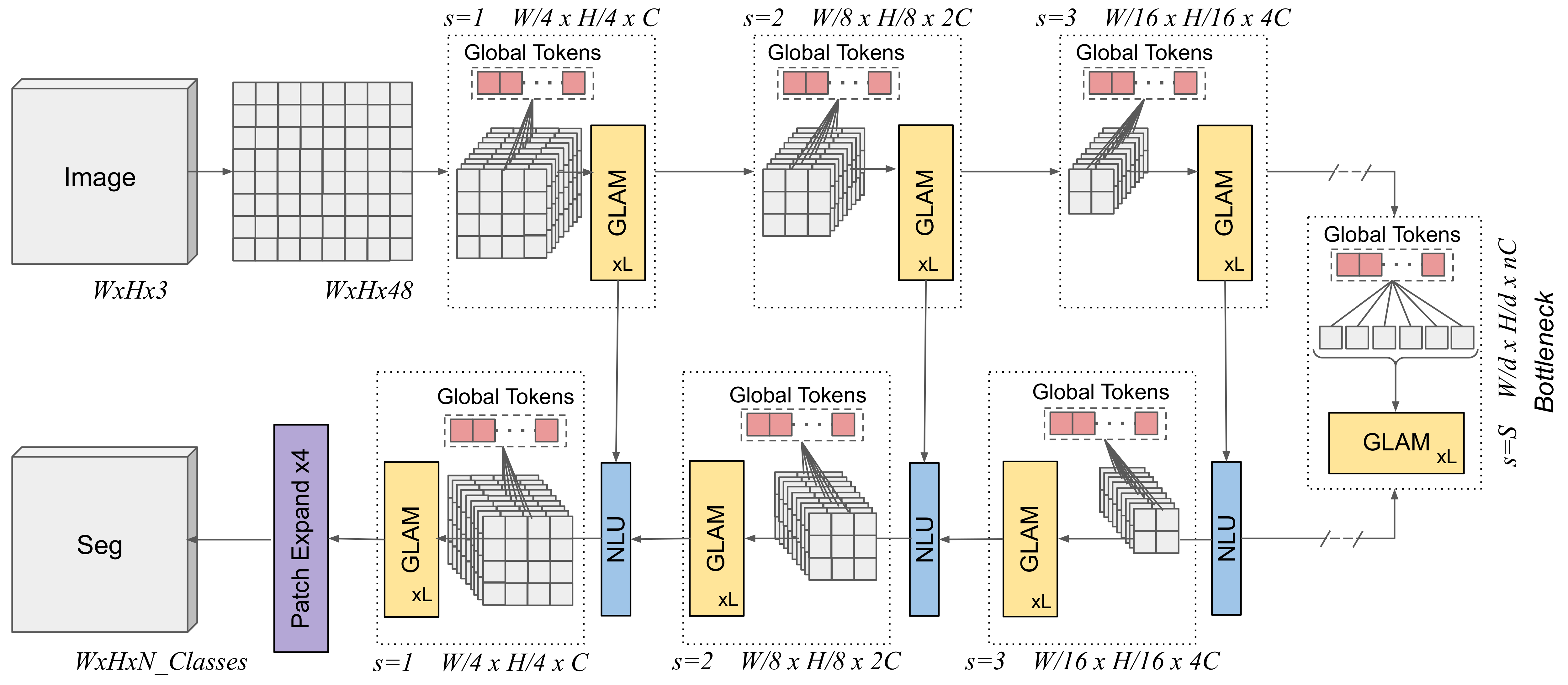}
    \caption{The GLAM module for modeling full-range interaction in multi-resolution transformers. GLAM is included at each resolution level of any multi-resolution transformer architecture, \eg Swin-Unet~\cite{liu2021Swin} or Swin-UperNet~\cite{liu2021Swin}. GLAM includes learnable global tokens, which are leveraged into a succession of two attention steps. We show that this design can indirectly represent long-range interactions between all image regions at all scales, and also external information useful for segmentation while retaining efficiency. We also introduce a non-local upsampling scheme (NLU) to extend the global context modeling in full transformer U-shape architectures such as \cite{cao2021swinunet,zhou2021nnformer}.}
     \label{fig:archi}
\end{figure*}

\section{Related work}
\textbf{Semantic Segmentation.} In the deep learning era, Fully Convolutional neural Networks (FCNs)~\cite{long_fully_2015,shelhamer2017fully_FCN,zhao2017pspnet,deeplabv3plus2018,xiao2018unified_upernet} have mainly led state-of-the-art performance in semantic segmentation. For example, DeepLab~\cite{deeplabv3plus2018} is a model based on an encoder-decoder architecture, and U-shape networks~\cite{ronneberger2015u} and 3D variants~\cite{milletari2016v,Isensee2020nnUNetAS} are extremely popular in medical image segmentation.
However, those models are limited to a local receptive field which is small for high-resolution images. Recently, transformers~\cite{AIAYN} have gained a lot of interest from their ability to model long-range interactions, which allows larger spatial context information to be exploited.

\noindent\textbf{Vision Transformer backbones.}
Building on the strong performances of transformers for auto-regressive inference, fully transformer-based models for image generation have been proposed~\cite{child2019generating}. Other early works proposed models combining CNNs and attention for vision tasks such as object detection~\cite{zhu2020deformable}, disparity estimation~\cite{li2021revisiting} or semantic segmentation~\cite{hou2020strip}.  More recently, fully transformer architectures have outperformed FCN baselines in image classification. ViT~\cite{dosovitskiy2020vit} is the first pure transformers backbone that achieved state-of-the-art performance for image classification but it requires very large training databases. DeiT~\cite{pmlr-v139-touvron21a} managed to reduce this requirement through data-efficient training strategies and distillation.

\noindent\textbf{Multi-resolution transformers.} Several recent approaches proposed adaptations of the vanilla ViT architecture. In particular, some architectures rely on multi-resolution processing. T2T ViT~\cite{Yuan_2021_ICCV} constructs richer semantic feature map through token aggregation while TnT~\cite{han2021TnT} and crossViT~\cite{chen2021crossvit} uses two transformers for fine and coarse resolution. PvT~\cite{wang2021pyramid} is the first backbone with a fully pyramidal architecture that is based on windowed transformers, allowing to process the images at fine resolution and to build rich feature maps, while reducing the spatial complexity. Other methods kept this hierarchical approach while improving information sharing between the windows. Swin~\cite{liu2021Swin} and its variant~\cite{cao2021swinunet,zhou2021nnformer} proposed to used shifted windows, Twins~\cite{chu2021Twins} uses interleaved fine and coarse resolution transformers, and CvT~\cite{wu2021cvt} replaces linear embedding with convolutions. 

\noindent\textbf{Efficient Self-Attention.} Long sequences have been a challenge for transformers because the original self-attention mechanism has a quadratic complexity in the sequence length. To tackle this, many approaches focus on designing efficient self-attention mechanisms.

Most of them are developed for NLP tasks and can be grouped into four categories. The first category uses a sparse approximation of the attention matrix~\cite{qiu2020blockwise,kitaev2019reformer,imagetransformer}. Among these approaches, window-based patch extraction vision transformers recently provided a simple yet efficient approach to compute attention~\cite{liu2021Swin,wang2021pvtv2,fan2021multiscale}. The second category is composed of methods based on a low-rank approximation of the attention matrix, such as Linformer~\cite{wang2020linformer}. The third category (memory-based transformers) construct buffers of extra tokens used as static memory~\cite{rae2019compressive,lee2019set}. The fourth category (kernel-based methods) provides a linear approximation of the softmax kernel~\cite{choromanski2020rethinking,peng2020random,katharopoulos2020transformers}. %Several approaches combine multiple efficient attention mechanisms such as Longformer~\cite{beltagy2020longformer} or ETC~\cite{ainslie2020etc}, using sparsity and memory mechanisms. Some combine them with other memory-efficient approaches such as Reformer~\cite{kitaev2019reformer} based on sparsity and reversibility. 
Some vision transformers have combined multiple efficient attention mechanisms. The recent ViT-inspired backbone PvT~\cite{wang2021pyramid} is based on windowed self-attention and attention approximation close to Linformer. ViL~\cite{zhang2021multi} balances sparse attention by using a reduced set of global tokens (usually a single one) to extract global representations of the input image.

The GLAM method introduced in this paper is a window- and memory-based transformer that fits well in existing multi-resolution vision backbones. Unlike most other multi-resolution backbones, GLAM fully extends the notion of windowed attention to semantic segmentation by introducing global tokens at the window level, and by designing a specific GLAM transformer cascading window (W-MSA) and global (G-MSA) attention.
%GLAM is related to ViL~\cite{zhang2021multi} by the use of global tokens, but fully extends the idea to semantic segmentation by introducing global tokens at the window level, and by designing a specific GLAM transformer cascading window (W-MSA) and global (G-MSA) attention. Moreover, ViL uses only one global token to retrieve information at the image level while GLAM uses multiple global tokens per window, allowing to grasp rich representations related to each image region.
 We highlight that GLAM enables global communication across all image regions and also encodes learned information from all the training sets.

\section{The GLAM Method}

The main idea in GLAM is to provide a way to represent full range interactions at all feature map resolutions, which is impossible in vanilla models, especially in high-resolution feature maps, due to the quadratic complexity of attention transformers. 

GLAM is illustrated in Fig.~\ref{fig:archi}, where it has been added to the Swin-Unet architecture~\cite{liu2021Swin}. Note that GLAM can be included in various multi-resolution architectures, \eg Swin~\cite{cao2021swinunet} or PvT~\cite{wang2021pyramid} and is also applicable for 3D segmentation, \eg nn-Former~\cite{zhou2021nnformer}. 
The core idea in GLAM is to design global tokens (in red in Fig.~\ref{fig:archi}), which are leveraged into a succession of two attention steps: first, between visual tokens in each window independently and, second, between global tokens among different windows. We show in Fig.~\ref{sec:GLAM} that this design enables to represent full range interactions between all image regions at all scales, and also external information useful for segmentation, while retaining efficiency. We also introduce a non-local upsampling scheme (NLU) to extend the full context modeling in U-shape architectures and to provide an efficient interpolation of rich semantic feature maps in an associated decoder.

\subsection{Multi-resolution transformer architecture}

As shown in Fig.~\ref{fig:archi}, GLAM can be included into any multi-resolution transformer architecture~\cite{liu2021Swin,wang2021pyramid,zhang2021aggregating,wang2021pvtv2,cao2021swinunet,zhou2021nnformer}.

\noindent\textbf{Transformer.} At each resolution level $s$, given a sequence of visual tokens, a transformer learns representations through Self-Attention (SA). 
SA is given by the expectation of each token with respect to their probability of sharing the same embedding. The Multi-head Self-Attention (MSA) is obtained from the linear combination of $m$ parallel SA operations. Finally, a complete transformer module is obtained by plugging the output of the MSA into a Multi-Layer Perceptron (MLP). 
Layer norm operations, as well as residual connections, are added respectively before and after MSA and MLP modules.

\noindent\textbf{Windowed attention.} MSA cannot be applied to long sequences \textit{e.g.}~patches from high-resolution images becasue the  computation of the attention matrix has quadratic memory complexity.
 To allow high-resolution processing and thus long sequences of small patches, windowed transformers treat the image as a batch of non-overlapping windows ~\cite{liu2021Swin,wang2021pyramid,wang2021pvtv2,zhang2021aggregating}. This approach is combined with a pooling strategy~\cite{cao2021swinunet,liu2021Swin,wang2021pyramid,zhang2021aggregating} and is well suited to build a multi-resolution encoder, able to produce rich semantic maps. Multi-resolution backbones are built by chaining windowed transformer blocks and downsampling.  These hierarchical architectures manage to build larger receptive fields in deeper layers, similarly to CNNs. This, however, does not guarantee a global receptive field and the maximal receptive field depends on the model's depth. More importantly, this processing introduces a major modification to the transformer modules. At a finer resolution, only local interactions are considered. With this modification, the processing of isolated patches by self-attention may not be as effective as global self-attention performed on the full image.

\subsection{Global attention multi-resolution transformers}
\label{sec:GLAM}
\begin{figure*}
    \centering
    \includegraphics[width=0.7\linewidth]{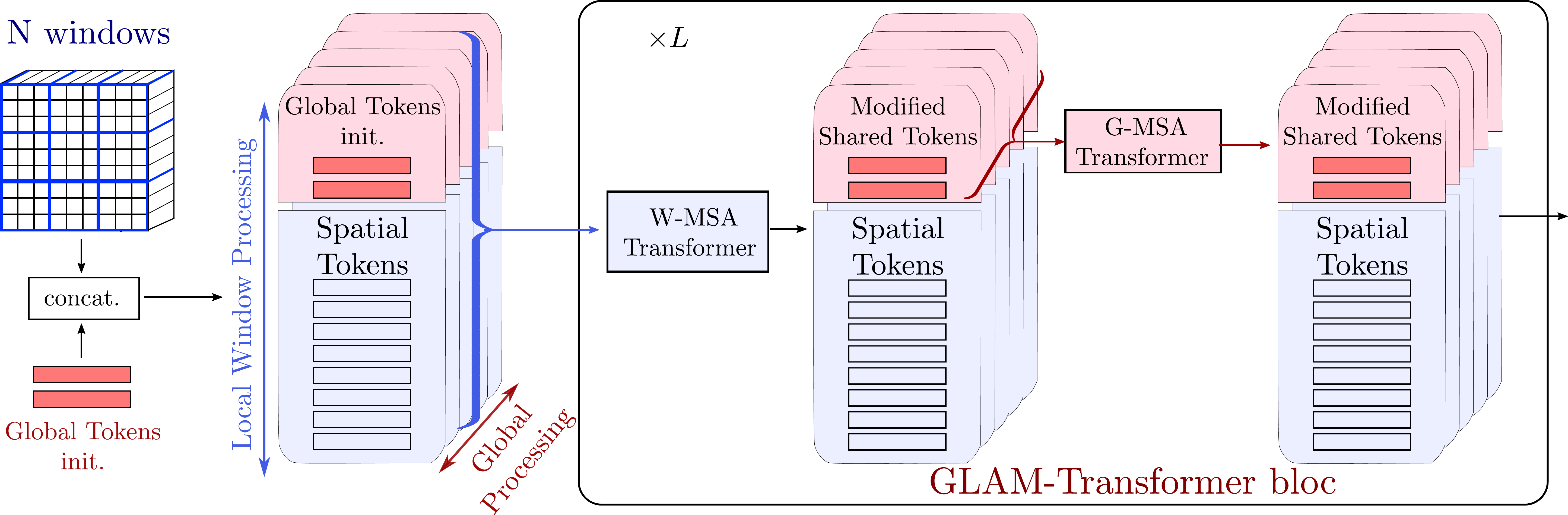}
    \caption{\textbf{GLAM-Transformer:} as in multi-resolution approaches, each input feature map is divided into $N_r$ non overlapping windows (blue). The core idea in GLAM is to design learnable global tokens (in red). The visual tokens from each window are concatenated with the global tokens and processed through a local window transformer (W-MSA). Every W-MSA is followed by a global transformer (G-MSA), where global tokens between different windows interact with each other, which brings a global representation to each window. These two steps give the GLAM-Transformer block; Multiple blocks are chained at every hierarchy level in typical multi-resolution transformer backbones. We show that global tokens learned from GLAM-Transformer indirectly model global interactions between all visual tokens in all widows. The global tokens are also able to represent extra learnable knowledge beyond the patch interactions in a single image. }
    \label{fig:gtokens}
\end{figure*}

We show how the GLAM module can provide global attention in all feature maps of multi-resolution transformers.
The GLAM transformer is illustrated in Fig.~\ref{fig:gtokens}, consisting in a sequence of $L$ transformer blocks, processing visual tokens in each region of the multi-resolution maps (shown in blue in Fig.~\ref{fig:gtokens}) and global tokens (shown in red in Fig.~\ref{fig:gtokens}). 

The basic idea behind GLAM is to associate global tokens to each window that are responsible to encapsulate the local information and transmit it to other image regions by computing MSA between all global tokens. Thus, when information is processed at the window scale, the visual tokens embedding incorporate useful long-range information.\\

\noindent\textbf{Global Tokens.} Global tokens lie at the core of Global Attention (GA). They are specific tokens concatenated to each window and are responsible for communication between windows.
We define as $N_r$ the number of windows in the feature map, $N_p$ as the number of patches per window, and $\{\V w_{k}^l\}_{1\leq k \leq N_r}$ as the sequence of windows after being processed by the $l^{th}$ GLAM-transformer block. We define as $\{\V g_k^{l}\}_{1\leq k \leq N_r}$ the sequence of $N_g$-dimensional global tokens associated to each window. The initialization of the global tokens $\{\V g_k^{0}\}_{1\leq k \leq N_r}$ is the same for all windows and is learned by the model. The input of the $l^{th}$ transformer block, defined as $\V z^l$, is a batch of tokens from each window concatenated with the corresponding global tokens, \textit{i.e.}~$\V z^l \in \mathbb R^{N_r\times (N_g + N_p)\times C}$ with $C$ being the dimension of the tokens. Consequently, the elements in the batch have the form:
\begin{align}
     &\forall k \in [1 \ldotp \ldotp N_r],  \V z_{k}^l = \begin{bmatrix} \V g_k^l \\[.25ex] \V w_k^l\end{bmatrix} \in \mathbb{R}^{(N_g + N_p)\times C}.
\end{align}

\noindent\textbf{GLAM-Transformer.} The communication between windows at a given hierarchy level is obtained through the interaction of global tokens. At each block $l$ of the GLAM-transformer, there are two steps: \textit{i)} visual tokens grasp their local statistics through a local window transformer (W-MSA), and \textit{ii)} the global tokens are re-embedded by a global transformer (G-MSA), where global tokens from different windows interact with each other. 
Formally, the $l^{th}$ GLAM-transformer block inputs $\V z^{l-1}$ and outputs $\V z^{l}$ by the succession of a W-MSA and  a G-MSA step:
\begin{align}
    &\Vh z^l = \text{W-MSA}(\V z^{l-1}) \,, \notag \\
    &\V g^l = \text{G-MSA}(\Vh g^{l}) \,, \notag \\
    &\V z^l = \begin{bmatrix} \V g_k^{l^T} \Vh w_k^{l^T} \end{bmatrix}^T
\end{align}
We define as $\V A_r^l$ the attention matrix for the window $r$ in the transformer block $l$. We introduce the following decomposition to express the attention with respect to the global and local tokens:
\begin{align}
    \V A_r^l = \begin{bmatrix}\V A_{r,gg}^l & \V A_{r,gw}^l \\ \V A_{r,wg}^l & \V A_{r,ww}^l \end{bmatrix} \, .
\end{align}

The square matrices $\V A_{r,gg}^l \in \mathbb{R}^{N_g \times N_g}$ and $\V A_{r,ww}^l \in \mathbb{R}^{N_p \times N_p}$ give the attention from the global token and the spatial tokens on themselves respectively. The matrices $\V A_{r,gw}^l \in \mathbb{R}^{N_g \times N_p}$ and $\V A_{r,wg}^l \in \mathbb{R}^{N_p \times N_g}$ are the cross attention matrices between local and global tokens. 
We define as $\V B^l \in \mathbb{R}^{(N_r \cdot N_g) \times (N_r \cdot N_g)}$ the global attention matrix from all the global token sequence and $\V B_{ij}^l \in \mathbb{R}^{N_g \times N_g}$ as the sub-matrices giving the attention between the global tokens of windows $i$ and $j$.

\noindent\textbf{GLAM-Transformer properties.} 
Putting aside the value matrix, the W-MSA gives the following embedding $\Vh g_r^l $ from $\V g_r^{l-1}$:
\begin{align}
    &\Vh g_r = \V A_{r,gg}^l \V g_r^{l-1} + \V A_{r,gw}^l \V w_r^{l-1} \, . 
\end{align}
The G-MSA, \textit{i.e.}~the MSA on the sequence of global tokens gives the following embeddings:
\begin{align}
    \V g_r^l &=\sum_{n=1}^{N_r} \V B_{rn}^l \Vh g_n^l \notag \\
    &=\sum_{n=1}^{N_r} \V B_{rn}^l (\V A_{r,gg}^l\V g_r^{l-1} + \V A_{r,gw}^l \V w_r^{l-1}) \, .
\label{eq:global_tokens}
\end{align}

From Eq.~(5), we have the expression of the global token  for a window $r$ processed by the $l$ G-MSA block transformer. Developing this formulation we obtain the following expression for the $k^{th}$ global token in the $r^{th}$ window:

\begin{align}
    g_{k,r}^l &= \sum_{r'=1}^{N_r} \sum_{j=1}^{N_g} b_{k,r,j,r'} \biggl( \sum_{i=1}^{N_g + N_p} a_{j, r', i} z^{l-1}_{i, r'} \biggr ) \notag \\
    &=\sum_{r'=1}^{N_r} \sum_{j=1}^{N_g} b_{k,r,j,r'} \biggl( \sum_{i=1}^{N_g} a_{j, r', i} g^{l-1}_{i, r'} \notag \\ &\qquad+ \sum_{i=1}^{N_p} a_{j, r', i + N_g} w^{l-1}_{i, r'} \biggr ).
\label{eq:detailed_ga}
\end{align}
The variables $z_{i,r}$, $g_{i,r}$ and $w_{i,r}$ corresponds respectively to the visual, global or generic token $i$ in window $r$. $a_{j,r,i}$ is the attention coefficient given by the token $j$ to the token $i$ inside the window $r$. $b_{j,r,i,r'}$ is the attention coefficient from the global token $j$ in the window $r$ to the global token $j$ in the window $r'$. Re-arranging the indices of equation \cref{eq:detailed_ga} leads to the following expression for the $k^{th}$ global token in the $r^{th}$ window:

\begin{align}
g_{k,r}^{l} & = 
 \sum\limits_{r'=1}^{N_r} \sum\limits_{i=1}^{N_p} \left(\sum\limits_{j=1}^{N_g} b_{k,r,j,r'} ~ a_{j,r',(i+N_g)}~w_{i,r'}^{l-1}\right) \nonumber \\
& + \sum\limits_{r'=1}^{N_r}\left(\sum\limits_{j=1}^{N_g} b_{k,r,j,r'} \sum\limits_{i=1}^{N_g}a_{j,r',i}~g_{i,r'}^{l-1}\right)  
\label{eq:lr-attention}
\end{align}

\noindent This leads to a global attention matrix $\V G_k \in \mathbb R^{(N_r \cdot N_p)\times(N_r \cdot N_p)}$ associated to the $k^{th}$ global token given by $[\V G_k]_{r',i} = \sum_{j=1}^{N_g} b_{k,r,j,r'}a_{j,r',(i+N_g)} +  \sum_{j=1}^{N_g} b_{k,r,j,r'} \sum_{i=1}^{N_g} a_{j,r',i} $. 
\cref{eq:lr-attention} gives the embedding of the the global token $g_{k,r}^{l}$ at the $l^{st}$ GLAM-transformer block, with respect to all visual tokens in all feature map windows $w_{i,r'}^{l-1}$ (first row), and all global tokens $g_{i,r'}^{l-1}$ (second row). This rewriting shows that the global embedding $g_{k,r}^{l}$ captures interactions between all image regions independently of the resolution. The different terms in the decomposition are interpreted as an attention map associated with each image region. This is the visualization shown in \cref{fig:intro}: the row of the first term corresponds to patch-based attention which depends on all the tokens of the feature map, while the second row represents window-based attention.

Overall, global tokens embedded with GLAM-transformers provide a way for information propagation across all windows (first row in Fig.~\ref{eq:lr-attention}), but also global information (second row) that goes beyond matching visual features in a single image. Especially, this represents global and learned information across the dataset, and can be leveraged as a stabilizing effect in SA, because the information is shared not only from the input but from all the windows in the dataset. This makes them a powerful tool to interpret isolated tokens and to take advantage of redundant structures in the data.

%\noindent\textbf{GLAM-Transformer complexity.} The computational complexity of an MSA module for an image $I$ divided into $h\times w$ patches has quadratic scaling with respect to the image area $h w$. The windowed approach W-MSA only depends on $N_p h w$. The complexity of both methods is given by:

%\begin{align}
%    &\Omega(\text{MSA}(I)) = 4hwc^2 + 2(hw)^2c \\
%    &\Omega(\text{W-MSA}(I)) = 4hwc^2 + 2N_p hwc
%\end{align}
%This makes the W-MSA scalable to a large number of patches where the MSA can not be computed. With few global tokens, the global attention adds only a few numbers of operations as it corresponds to adding $N_g$ tokens in each window and performing MSA over a sequence of length $N_g \times N_r$. 
%It is also worth noting that the global tokens add a limited memory overhead as they do not require any more activation saving and only add a few elements in the attention matrix from each transformer block. 

\noindent\textbf{Non-Local Upsampling.} 
We introduce a Non-Local Upsampling (NLU) module for a fully transformer decoder such as \cite{cao2021swinunet,zhou2021nnformer}. NLU is designed to upsample the semantic features based on all the tokens coming from the skip connection, by drawing inspiration for non-local means~\cite{Buades:2005:NAI:1068508.1069066}. 

The proposed NLU is illustrated in the supplementary material. To perform the upsampling, the skip connections are embedded into a query matrix of size ${(4 N_p)\times C}$ while the semantic low-resolution features are embedded into the keys and values of size ${N_p\times C}$. The projection of the values on the resulting attention matrix has the size ${(4N_p)\times C}$.

\section{Experiments}

\subsection{Experimental Settings}

\noindent\textbf{Datasets.} We evaluated on three different semantic segmentation datasets: ADE20K~\cite{ADE20K}, Cityscapes~\cite{Cityscapes} and Synapse~\cite{synapse}. ADE20K is a scene parsing dataset composed of 20,210 images with 150 object classes. Cityscapes contains driving scenes and is composed of 5,000 images annotated with 19 different classes. Synapse is an abdominal organ segmentation dataset that includes 30 Computerized Tomography (CT) scans which are 3D volumes annotated with 8 abdominal organs.

\noindent\textbf{Implementation details.} GLAM models were implemented into the mmseg~\cite{mmseg2020} codebase and the models were trained on 8 Tesla V100 GPUs. The layers were pretrained on ImageNet-1K and standard augmentation was used: random crop, rotations, translations, \textit{etc.} More details are provided in supplementary. We used the Adam optimizer with a weigh decay of 0.01 and a polynomial learning rate scheduler starting from 0.00006 and with a factor of 1.0.  The reported segmentation performances are mean Intersection over Union (mIoU) for ADE20k and Cityscapes and Dice Similarity Score (DSC) for Synapse. 

\subsection{GLAM performance}

\begin{table}[h!]
    \centering
    \caption{\textbf{GLAM Improvements on  various multi-resolution transformers.} Performances are evaluated with respect to mIoU for ADE20k and Cityscapes and average DSC for Synapse.}
    \label{tab:swinlike}
   % \resizebox{\columnwidth}{!}{%
    \begin{tabular}{c|l|cc}
        \toprule[1pt]
        Dataset & Method &  Size  & Score        \\
        \hline
        & Swin-Unet~\cite{cao2021swinunet} & Tiny & 42.75   \\
        & GLAM-Swin-Unet & Tiny  & \textbf{44.19}    \\
         \cline{2-4}
        & Swin-UNet~\cite{cao2021swinunet} & Small & 47.49   \\
        & GLAM-Swin-UNet & Small  & \textbf{47.90}    \\ 
        \cline{2-4}
        & Swin-Unet~\cite{cao2021swinunet}  & Base   &   47.85   \\
         & GLAM-Swin-Unet & Base  & \textbf{49.10} \\
        \cline{2-4}
        & Swin-UperNet\cite{liu2021Swin} & Tiny &     43.69      \\
        & GLAM-Swin-UperNet & Tiny &  \textbf{44.16}     \\%\textbf{43.93}     \\
         \cline{2-4}
        & Swin-UperNet~\cite{liu2021Swin} & Small & 47.72   \\
        & GLAM-Swin-UperNet & Small  & \textbf{47.75}    \\
        \cline{2-4}
        & Swin-UperNet~\cite{liu2021Swin} & Base &     47.99           \\
        \multirow{-10}{*}[-3ex]{ADE20K} & GLAM-Swin-UperNet & Base & \textbf{48.44} \\        
        \midrule[1pt]
        & Swin-UperNet~\cite{liu2021Swin} & Tiny &  78.24           \\
        & GLAM-Swin-UperNet & Tiny & \textbf{78.64} \\
        \cline{2-4}
        & Swin-UperNet~\cite{liu2021Swin} & Base &  80.79           \\
        & GLAM-Swin-UperNet & Base & \textbf{81.47} \\
        \cline{2-4}
        & Swin-Unet~\cite{cao2021swinunet} & Tiny & 77.43   \\
        \multirow{-4}{*}{Cityscapes} & GLAM-Swin-Unet & Tiny & \textbf{78.29}    \\
        \midrule[1pt]
        & nnFormer~\cite{zhou2021nnformer} & Tiny & 87.40   \\
        \multirow{-2}{*}{Synapse} & GLAM-nnFormer & Tiny & \textbf{88.60}    \\
        \bottomrule[1pt]
    \end{tabular}
    %}
    \vspace{-0.3cm}
\end{table}
\noindent\textbf{GLAM in multi-resolution transformers.}
GLAM is well suited to work with window transformers such as PvT~\cite{wang2021pyramid,wang2021pvtv2} or Swin~\cite{liu2021Swin} as well as its variants~\cite{cao2021swinunet,zhou2021nnformer}. Due to the top performances of Swin, we incorporated GLAM into this backbone to compute the segmentation of 2D datasets leading to two models: GLAM-Swin-UperNet and GLAM-Swin-Unet. The first one is a hybrid model combining a transformer backbone and a CNN head~\cite{cao2021swinunet,xiao2018unified_upernet} while the second one is a full transformer model with a decoder symmetric to the encoder~\cite{cao2021swinunet}. For 3D images, GLAM was plugged into nnFormer~\cite{zhou2021nnformer} which is designed similarly to Swin-Unet for 3D medical image segmentation. The performances of the Swin and GLAM models are presented in Table~\ref{tab:swinlike}. GLAM models exhibit important and consistent performance gains compared to their vanilla counterparts, either on small or larger models: \eg $\sim$ +1.5pt gain on ADE20K with Swin-Unet (Base or Tiny), and +1.2pt on Synapse on the recent nn-Former model.

\noindent\textbf{State-of-the-art comparison.} We now compare the GLAM-Swin models with existing approaches on the ADE20K~\cite{ADE20K}, Cityscapes~\cite{Cityscapes} and Synapse~\cite{synapse}.\\ 

\noindent\textbf{ADE20K and Cityscapes.} Table~\ref{tab:sota_ade_city} summarizes our results. To be fair, we compared models up to $\sim$ 150M parameters, and we report the top performances from the mmseg~\cite{mmseg2020} benchmark for all methods, with 160K training epochs for all methods. Moreover, we compared only methods trained on $768 \times 768$ resolution images on Cityscapes. In this setup, GLAM-Swin-Unet yields 49.10\% mIoU on ADE20K outperforming its vanilla Swin counterpart with at least 1.10\% mIoU. GLAM-Swin-UperNet achieves 81.47 \% mIoU on Cityscapes which is 1.58 \% better than its Swin-Upernet counterpart.  %SegFormer~\cite{xie2021segformer} is the only very recent method achieving better performances on both datasets in the considered setup, but this gain essentially comes from a stronger model pre-training. SegFormer's low-resolution attention could however be replaced with a GLAM module to further improve performances. 
\\

\noindent\textbf{Synapse.} Table~\ref{tab:sota_syna} reports our results and recent baselines for 3D medical segmentation. GLAM-nnFormer significantly outperforms all other existing methods by at least 1.2\% average Dice. To the best of our knowledge, GLAM-nnFormer outperforms state-of-the-art on the Synapse dataset.

\begin{table}[h!]
    \centering
    \caption{Comparison to state of the art methods on ADE20K and Cityscapes. All experiments are made or reported are with single-scale inference.}
    \label{tab:sota_ade_city}
    \resizebox{\linewidth}{!}{%
    \begin{tabular}{l|lc|c}
        \toprule[1pt]
        & & \multicolumn{1}{c}{ADE20K} & \multicolumn{1}{c}{Cityscapes}\\
        Method & Backone & mIoU & mIoU \\
        \hline
        FCN \cite{shelhamer2017fully_FCN}         & ResNet-101      & 41.40 & 77.34\\
        CCNet \cite{huang2018ccnet}       & ResNet-101        & 43.71 & 79.45\\
        %OCRNet \cite{YuanCW20_OCRNet}      & HRNet       	    & 43.25 & 80.58\\
        DANet \cite{fu2018dual_DANet}        & ResNet-101  	   & 43.64 & 80.47\\
        UperNet \cite{xiao2018unified_upernet}     & ResNet-101      & 43.82 & 80.10\\
        DNL \cite{yin2020disentangled_DNL} & ResNet-101      & 44.25 & 79.41\\
        PSPNet \cite{zhao2017pspnet}  & ResNet-101      & 44.39 & 79.08\\
        DeepLabV3+ \cite{deeplabv3plus2018}  & ResNet-101      & 45.47 & 79.41\\
        \hline 
        Trans2Seg  \cite{wang2021pyramid}  & PVT-S           & 42.60 & - \\
        %Swin-Unet    & Swin-T          &        &      \\
        FPN \cite{wang2021pyramid}          & PVT-L           & 42.10 & -\\
        TNT \cite{han2021TnT} & TNT-S & 43.60 & - \\
        SETR-PUP \cite{SETR} & DeiT-L           & 46.34 & 79.21\\
        Swin-Unet \cite{cao2021swinunet}   & Swin-B           &   47.85 & - \\
        Swin-UperNet  \cite{liu2021Swin}    & Swin-B          & 47.99 & 80.79\\ 
        Twins-SVT-L \cite{chu2021Twins} & Twins-SVT & 48.80 & - \\
        %SegFormer \cite{xie2021segformer}      & MiT-B5 &\textbf{50.36} & \textbf{82.3} \\
        \hline
        GLAM-Swin-Unet & Swin-B & {\textbf{49.10}} & - \\ % Swin-Unet base + 10 GT
        GLAM-Swin-UperNet  & Swin-B & 48.44 & {\textbf{81.47}}\\ % Swin-Unet base + 10 GT
        \toprule[1pt]
    \end{tabular}
    }
    %\vspace{-0.1cm}
\end{table}

\begin{table}[h!]
    \centering
    \caption{Comparison to state of the art methods on Synapse.}%(Dice Score in \%)
    \label{tab:sota_syna}
    %\resizebox{\linewidth}{!}{%
    \begin{tabularx}{\linewidth}{X|>{\centering\arraybackslash}X}
        \toprule[1pt]
        Methods & Average Dice Score (\%) \\
        \hline
        VNet \cite{milletari2016v} & 68.81  \\
        U-Net \cite{ronneberger2015u} & 76.85  \\
        Att-UNet \cite{oktay2018attention} & 77.77  \\
        R50-Deeplabv3+ \cite{deeplabv3plus2018} & 75.73 \\
        TransUNet \cite{chen2021transunet} & 77.48 \\
        Swin-Unet \cite{cao2021swinunet} & 79.13 \\
        TransClaw U-Net \cite{chang2021transclaw} & 78.09 \\
        nnUNet (3D) \cite{Isensee2020nnUNetAS} & 86.99  \\
        nnFormer \cite{zhou2021nnformer} & 87.40  \\
        GLAM-nnFormer & \textbf{88.60} \\
        \bottomrule[1pt]
    \end{tabularx}
    %}
    \vspace{-0.1cm}
\end{table}

\subsection{Model Analysis}

In this part, we analyze various important aspects of GLAM.

\begin{figure}[h!]
    \centering
    \includegraphics[width=0.7\linewidth]{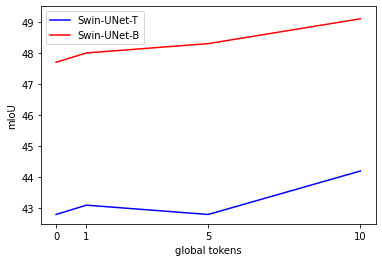}
    \caption{Impact of the number of global tokens on performance (mIoU) using ADE20k.}
    \label{fig:gt_impact}
\end{figure}

\textbf{Number of Global Tokens.}
The number of global tokens directly influences the capacity of GLAM to model global interactions between the windows.
Fig.~\ref{fig:gt_impact} shows the impact of this hyper-parameter on segmentation performances. 
 We can see that using more global tokens improves performances. However, it also increases the number of parameters and memory cost which forces a trade-off. We keep a reasonable value of 10 global tokens, which gives an important performance boost of +1.4pts in both the tiny and base versions of the Swin-Unet model.

\noindent\textbf{Impact of NLU.}
GLAM improves context modeling in multi-resolution transformers thanks to global attention and Non-Local Upsampling (NLU).
Table~\ref{tab:ablation_ga_nlu} provides an ablation study of these two components.
We can see that NLU gives an improvement 0.45pt compared to the original Swin-Unet that uses a patch expension operation for upsampling. GLAM brings another large improvement for a total gain of +1.44pts compared to the baseline.

\noindent\textbf{Long-range interaction.} To highlight the impact of G-MSA, Table \ref{tab:ablation_gmsa} shows the performances of GLAM backbones using only a W-MSA step but no G-MSA. GLAM backbones show consistent gains compared to their counterparts without G-MSA. This ablation highlight the crucial role of this step to leverage long-range interactions and that the performance gains made by GLAM can not only be explained by the parameter overhead. 
\begin{figure*}[!h]
\begin{center}
    \begin{tabular}{c}
    \includegraphics[width=0.9\linewidth]{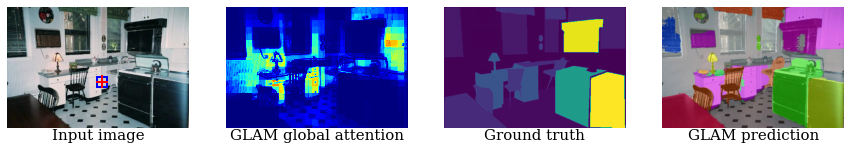}  \\
    a) Segmentation results and global attention of GLAM on ADE20K. \\
    \includegraphics[width=0.9\linewidth]{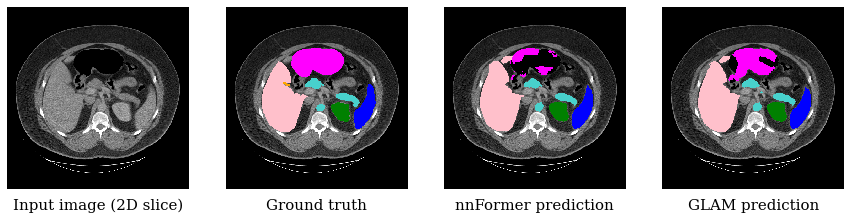}  \\
    b) Segmentation results on Synapse.
     \end{tabular}
      \caption{Qualitative visualisations of GLAM. We show the ability of GLAM to model full contextual information in high-resolution feature maps on ADE20K (first row), and the ability of GLAM-nn-Former to accurately segment the stomach (in pink).}
      \label{fig:visu_all}
      %\vspace{-1.9cm}
\end{center}      
\end{figure*}

\begin{table}
%\begin{minipage}{.24\linewidth}
%\begin{table}[h!]
    \centering
    \caption{Impact of the NLU and the GLAM transformer on a tiny Swin-Unet, 10 global tokens, on ADE20k.}
    \label{tab:ablation_ga_nlu}
    \begin{tabular}{c|cc|c}
        \toprule[1pt]
        Method & NLU & GLAM  & mIoU \\
        \hline
        Swin-Unet-T   &   &   & 42.75 \\
        Swin-Unet-T   & $\checkmark$ &   & 43.20 \\
        Swin-Unet-T   & $\checkmark$ & $\checkmark$ & \textbf{44.20} \\
        \toprule[1pt]
    \end{tabular}
    \vspace{-0.3cm}
%\end{table}
%\end{minipage}%%
\end{table}

\begin{table}[h!]
%\begin{minipage}{.24\linewidth}
%\begin{table}[h!]
    \centering
    \caption{Impact of G-MSA phase on GLAM transformer on different model, 10 global tokens, on ADE20k. GLAM-nogmsa is GLAM without the G-MSA phase.}
    \label{tab:ablation_gmsa}
    \begin{tabular}{l|c}
        \toprule[1pt]
        Method   & mIoU \\
        \hline
        %GLAM-Swin-Unet-T    &              & 43.60 \\
        %GLAM-Swin-Unet-T    & $\checkmark$ & \textbf{44.19} \\
        %\hline
        GLAM-nogmsa-Swin-Unet B    & 47.90 \\
        GLAM-Swin-Unet B    & \textbf{49.10} \\
        % \hline
        % GLAM-Swin-UperNet-T &              & \textbf{44.38} \\
        % GLAM-Swin-UperNet-T & $\checkmark$ & 44.16 \\
        % \hline
        GLAM-nogmsa-Swin-UperNet B &  47.95  \\
        GLAM-Swin-UperNet B &  \textbf{48.44} \\
        \toprule[1pt]
    \end{tabular}
    \vspace{-0.3cm}
%\end{table}
%\end{minipage}%%
\end{table}

\noindent\textbf{Parameter and FLOPs overhead.}
The overhead due to the global tokens is controlled and proportional to the number of GLAM transformer blocks. This overhead brings higher performance gains than increasing the backbone size which validates the model architecture.
\cref{tab:params} illustrates that the GLAM-Swin Base backbones show superior efficiency compared to their vanilla Large counterpart with a superior mIoU increase with respect to additional learnable parameters. The same analysis can be done with FLOPs overhead with a higher mIoU increase per extra-FLOP for GLAM-Swin Base compared to Swin Large. %Extended results on parameter overhead analysis is given in the supplementary material.

%\noindent\textbf{FLOPs overhead.} The computational overhead due to GLAM is controlled. Indeed, \cref{tab:params} illustrates that mIoU increase per extra-FLOP of GLAM-Swin Base shows superior efficiency compared to its Large counterparts.

%\begin{minipage}{\linewidth}
\begin{table}[h]
   \centering
   \caption{Analysis of the relative mIoU increase with respect to extra learnable parameters and FLOPs compared to the standard Base and Large backbones.}
   \resizebox{\linewidth}{!}{%
    \begin{tabular}{l|cc|cc}
    \toprule
backbone &  \#param. & \thead{$\uparrow$ rel. mIoU / \#param \\ $\times 10^{-2}$} & FLOPs & \thead{$\uparrow$ rel. mIoU / FLOPs\\ $\times 10^{-2}$} \\ 
\midrule
Swin-UperNet B      & 121 & 0            &   81G  & 0   \\
Swin-UperNet L      & 234 & 0.4          &   180G & 0.4 \\
\thead{ GLAM-Swin-\\UperNet B} & 197 & \textbf{0.6} &   99G  & \textbf{2.5} \\
    \bottomrule
    \end{tabular}
}
    \label{tab:params}
\end{table}
%\end{minipage}

\begin{figure}[h!]
    \centering
    \includegraphics[width=0.8\linewidth]{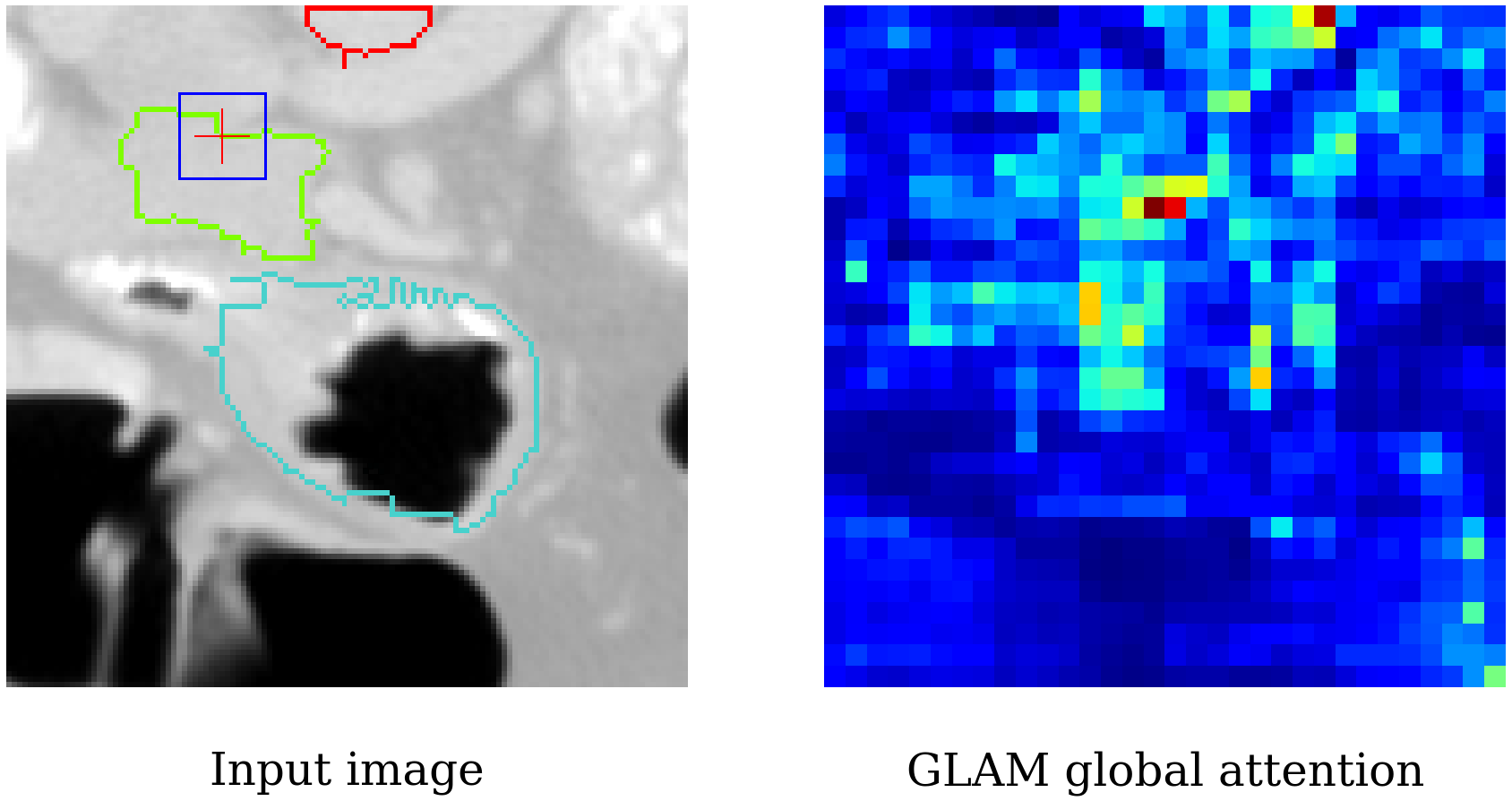}
    \caption{\textbf{Averaged GLAM attention map in 3D.} The information inside the blue window is ambiguous. To segment the voxel at the red cross, the model leverages long-range dependencies including neighbor organs. The pancreas is in green, the aorta in red, and the stomach in blue.\vspace{-1.5ex}}
    \label{fig:med_att}
\end{figure}

\noindent\textbf{Visualizations.}
Fig.~\cref{fig:visu_all} shows qualitative visualizations of the GLAM method. In Fig.~\cref{fig:visu_all}a), we show GLAM attention maps for the highest resolution feature maps of a GLAM Swin-Unet model. Echoing observations in Fig.~\cref{fig:intro} in Cityscape, we can see that GLAM can model full-range interactions in this spatially-fine layer. This enables to exploit spatial relationships with other important structures (\eg other sofas, arcades), which is not possible with the baseline Swin-Unet due to its limited window attention. We can notice the relevance of the GLAM segmentation. Furthermore, Fig.~\cref{fig:med_att} shows the GLAM attention averaged over the axial direction for the red cross (pancreas). We can see that long-range dependencies are involved, with a much larger spatial extent than the local window (in blue), where attention is given to neighboring organs (stomach and aorta). The full context is crucial to properly segment complex organs with visual local ambiguities such as the pancreas.
In Fig.~\cref{fig:visu_all}b), we show segmentation results of GLAM-nn-Former for 3D medical image segmentation. We show the results on a given 2D slice. We can notice that GLAM nn-Former is qualitatively much better at segmenting the stomach (in pink) than nn-Former. This can be explained by the global interactions of our model, which enables it to better represent specific interactions between organs.

\section{Conclusion}

This paper introduces GLAM, a method for modeling full contextual interactions in multi-resolution transformer-based models. the GLAM transformer leverage learnable global tokens at each resolution level of the model, which allows a complete interaction of the tokens across the image regions, and is further equipped with a non-local upsampling module. Experiments show the large and consistent gain of GLAM when incorporated into several multi-resolution transformers (Swin-Unet, nn-Former, Swin) on diverse medical, street, or more general images. Future works includes applying the GLAM idea for modeling full contextual information on very high-resolution images or 3D medical volumes.

{\small
\bibliographystyle{ieee_fullname}
\bibliography{egbib}

\begin{thebibliography}{10}\itemsep=-1pt

\bibitem{Buades:2005:NAI:1068508.1069066}
Antoni Buades, Bartomeu Coll, and Jean-Michel Morel.
\newblock A non-local algorithm for image denoising.
\newblock In {\em Proceedings of the 2005 IEEE Computer Society Conference on
  Computer Vision and Pattern Recognition (CVPR'05) - Volume 2 - Volume 02},
  CVPR '05, pages 60--65, Washington, DC, USA, 2005. IEEE Computer Society.

\bibitem{cao2021swinunet}
Hu Cao, Yueyue Wang, Joy Chen, Dongsheng Jiang, Xiaopeng Zhang, Qi Tian, and
  Manning Wang.
\newblock Swin-unet: Unet-like pure transformer for medical image segmentation,
  2021.

\bibitem{chang2021transclaw}
Yao Chang, Hu Menghan, Zhai Guangtao, and Zhang Xiao-Ping.
\newblock Transclaw u-net: Claw u-net with transformers for medical image
  segmentation, 2021.

\bibitem{chen2021crossvit}
Chun-Fu Chen, Quanfu Fan, and Rameswar Panda.
\newblock Crossvit: Cross-attention multi-scale vision transformer for image
  classification.
\newblock 2021.

\bibitem{chen2021transunet}
Jieneng Chen, Yongyi Lu, Qihang Yu, Xiangde Luo, Ehsan Adeli, Yan Wang, Le Lu,
  Alan~L. Yuille, and Yuyin Zhou.
\newblock Transunet: Transformers make strong encoders for medical image
  segmentation.
\newblock {\em arXiv preprint arXiv:2102.04306}, 2021.

\bibitem{deeplabv3plus2018}
Liang-Chieh Chen, Yukun Zhu, George Papandreou, Florian Schroff, and Hartwig
  Adam.
\newblock Encoder-decoder with atrous separable convolution for semantic image
  segmentation.
\newblock In {\em ECCV}, 2018.

\bibitem{child2019generating}
Rewon Child, Scott Gray, Alec Radford, and Ilya Sutskever.
\newblock Generating long sequences with sparse transformers.
\newblock {\em arXiv preprint arXiv:1904.10509}, 2019.

\bibitem{choromanski2020rethinking}
Krzysztof~Marcin Choromanski, Valerii Likhosherstov, David Dohan, Xingyou Song,
  Andreea Gane, Tamas Sarlos, Peter Hawkins, Jared~Quincy Davis, Afroz
  Mohiuddin, Lukasz Kaiser, et~al.
\newblock Rethinking attention with performers.
\newblock In {\em International Conference on Learning Representations}, 2020.

\bibitem{chu2021Twins}
Xiangxiang Chu, Zhi Tian, Yuqing Wang, Bo Zhang, Haibing Ren, Xiaolin Wei,
  Huaxia Xia, and Chunhua Shen.
\newblock Twins: Revisiting the design of spatial attention in vision
  transformers.
\newblock In {\em NeurIPS 2021}, 2021.

\bibitem{mmseg2020}
MMSegmentation Contributors.
\newblock {MMSegmentation}: Openmmlab semantic segmentation toolbox and
  benchmark.
\newblock \url{https://github.com/open-mmlab/mmsegmentation}, 2020.

\bibitem{Cityscapes}
Marius Cordts, Mohamed Omran, Sebastian Ramos, Timo Rehfeld, Markus Enzweiler,
  Rodrigo Benenson, Uwe Franke, Stefan Roth, and Bernt Schiele.
\newblock The cityscapes dataset for semantic urban scene understanding.
\newblock In {\em Proceedings of the IEEE conference on computer vision and
  pattern recognition}, pages 3213--3223, 2016.

\bibitem{dosovitskiy2020vit}
Alexey Dosovitskiy, Lucas Beyer, Alexander Kolesnikov, Dirk Weissenborn,
  Xiaohua Zhai, Thomas Unterthiner, Mostafa Dehghani, Matthias Minderer, Georg
  Heigold, Sylvain Gelly, Jakob Uszkoreit, and Neil Houlsby.
\newblock An image is worth 16x16 words: Transformers for image recognition at
  scale.
\newblock {\em ICLR}, 2021.

\bibitem{fan2021multiscale}
Haoqi Fan, Bo Xiong, Karttikeya Mangalam, Yanghao Li, Zhicheng Yan, Jitendra
  Malik, and Christoph Feichtenhofer.
\newblock Multiscale vision transformers.
\newblock {\em arXiv preprint arXiv:2104.11227}, 2021.

\bibitem{fu2018dual_DANet}
Jun Fu, Jing Liu, Haijie Tian, Yong Li, Yongjun Bao, Zhiwei Fang, and Hanqing
  Lu.
\newblock Dual attention network for scene segmentation.
\newblock 2019.

\bibitem{han2021TnT}
Kai Han, An Xiao, Enhua Wu, Jianyuan Guo, Chunjing Xu, and Yunhe Wang.
\newblock Transformer in transformer, 2021.

\bibitem{hou2020strip}
Qibin Hou, Li Zhang, Ming-Ming Cheng, and Jiashi Feng.
\newblock Strip pooling: Rethinking spatial pooling for scene parsing.
\newblock In {\em Proceedings of the IEEE/CVF Conference on Computer Vision and
  Pattern Recognition}, pages 4003--4012, 2020.

\bibitem{huang2018ccnet}
Zilong Huang, Xinggang Wang, Lichao Huang, Chang Huang, Yunchao Wei, and Wenyu
  Liu.
\newblock Ccnet: Criss-cross attention for semantic segmentation.
\newblock 2019.

\bibitem{Isensee2020nnUNetAS}
Fabian Isensee, Paul~F. Jaeger, Simon A.~A. Kohl, Jens Petersen, and Klaus
  Maier-Hein.
\newblock nnu-net: a self-configuring method for deep learning-based biomedical
  image segmentation.
\newblock {\em Nature methods}, 2020.

\bibitem{katharopoulos2020transformers}
Angelos Katharopoulos, Apoorv Vyas, Nikolaos Pappas, and Fran{\c{c}}ois
  Fleuret.
\newblock Transformers are rnns: Fast autoregressive transformers with linear
  attention.
\newblock 2020.

\bibitem{kitaev2019reformer}
Nikita Kitaev, Lukasz Kaiser, and Anselm Levskaya.
\newblock Reformer: The efficient transformer.
\newblock In {\em International Conference on Learning Representations}, 2019.

\bibitem{synapse}
Bennett Landman, Zhoubing Xu, Igelsias Eugenio, Juan, Martin Styner, Thomas
  Robin, Langerak, and Arno Klein.
\newblock Multi-atlas labeling beyond the cranial vault.
\newblock {\em MICCAI}, 2015.

\bibitem{lee2019set}
Juho Lee, Yoonho Lee, Jungtaek Kim, Adam Kosiorek, Seungjin Choi, and Yee~Whye
  Teh.
\newblock Set transformer: A framework for attention-based
  permutation-invariant neural networks.
\newblock In {\em Proceedings of the 36th International Conference on Machine
  Learning}, pages 3744--3753, 2019.

\bibitem{li2021revisiting}
Zhaoshuo Li, Xingtong Liu, Nathan Drenkow, Andy Ding, Francis~X. Creighton,
  Russell~H. Taylor, and Mathias Unberath.
\newblock Revisiting stereo depth estimation from a sequence-to-sequence
  perspective with transformers, 2021.

\bibitem{liu2021Swin}
Ze Liu, Yutong Lin, Yue Cao, Han Hu, Yixuan Wei, Zheng Zhang, Stephen Lin, and
  Baining Guo.
\newblock Swin transformer: Hierarchical vision transformer using shifted
  windows.
\newblock {\em International Conference on Computer Vision (ICCV)}, 2021.

\bibitem{long_fully_2015}
Jonathan Long, Evan Shelhamer, and Trevor Darrell.
\newblock Fully convolutional networks for semantic segmentation.
\newblock pages 3431--3440.

\bibitem{milletari2016v}
Fausto Milletari, Nassir Navab, and Seyed-Ahmad Ahmadi.
\newblock V-net: Fully convolutional neural networks for volumetric medical
  image segmentation.
\newblock In {\em 2016 fourth international conference on 3D vision (3DV)},
  pages 565--571. IEEE, 2016.

\bibitem{oktay2018attention}
Ozan Oktay, Jo Schlemper, Loic~Le Folgoc, Matthew Lee, Mattias Heinrich,
  Kazunari Misawa, Kensaku Mori, Steven McDonagh, Nils~Y Hammerla, Bernhard
  Kainz, et~al.
\newblock Attention u-net: Learning where to look for the pancreas.
\newblock {\em arXiv preprint arXiv:1804.03999}, 2018.

\bibitem{imagetransformer}
Niki~J. Parmar, Ashish Vaswani, Jakob Uszkoreit, Lukasz Kaiser, Noam Shazeer,
  Alexander Ku, and Dustin Tran.
\newblock Image transformer.
\newblock In {\em International Conference on Machine Learning (ICML)}, 2018.

\bibitem{peng2020random}
Hao Peng, Nikolaos Pappas, Dani Yogatama, Roy Schwartz, Noah Smith, and
  Lingpeng Kong.
\newblock Random feature attention.
\newblock In {\em International Conference on Learning Representations}, 2020.

\bibitem{qiu2020blockwise}
Jiezhong Qiu, Hao Ma, Omer Levy, Scott~Wen tau Yih, Sinong Wang, and Jie Tang.
\newblock Blockwise self-attention for long document understanding, 2020.

\bibitem{rae2019compressive}
Jack~W Rae, Anna Potapenko, Siddhant~M Jayakumar, Chloe Hillier, and Timothy~P
  Lillicrap.
\newblock Compressive transformers for long-range sequence modelling.
\newblock In {\em International Conference on Learning Representations}, 2019.

\bibitem{ronneberger2015u}
Olaf Ronneberger, Philipp Fischer, and Thomas Brox.
\newblock U-net: Convolutional networks for biomedical image segmentation.
\newblock In {\em International Conference on Medical image computing and
  computer-assisted intervention}, pages 234--241. Springer, 2015.

\bibitem{shelhamer2017fully_FCN}
Evan Shelhamer, Jonathan Long, and Trevor Darrell.
\newblock Fully convolutional networks for semantic segmentation.
\newblock {\em IEEE transactions on pattern analysis and machine intelligence},
  39(4):640--651, 2017.

\bibitem{pmlr-v139-touvron21a}
Hugo Touvron, Matthieu Cord, Matthijs Douze, Francisco Massa, Alexandre
  Sablayrolles, and Herve Jegou.
\newblock Training data-efficient image transformers \&; distillation through
  attention.
\newblock In {\em International Conference on Machine Learning}, volume 139,
  pages 10347--10357, July 2021.

\bibitem{AIAYN}
Ashish Vaswani, Noam Shazeer, Niki Parmar, Jakob Uszkoreit, Llion Jones,
  Aidan~N Gomez, \L~ukasz Kaiser, and Illia Polosukhin.
\newblock Attention is all you need.
\newblock In I. Guyon, U.~V. Luxburg, S. Bengio, H. Wallach, R. Fergus, S.
  Vishwanathan, and R. Garnett, editors, {\em Advances in Neural Information
  Processing Systems}, volume~30. Curran Associates, Inc., 2017.

\bibitem{wang2020linformer}
Sinong Wang, Belinda~Z Li, Madian Khabsa, Han Fang, and Hao Ma.
\newblock Linformer: Self-attention with linear complexity.
\newblock {\em arXiv e-prints}, pages arXiv--2006, 2020.

\bibitem{wang2021pvtv2}
Wenhai Wang, Enze Xie, Xiang Li, Deng-Ping Fan, Kaitao Song, Ding Liang, Tong
  Lu, Ping Luo, and Ling Shao.
\newblock Pvtv2: Improved baselines with pyramid vision transformer.
\newblock {\em arXiv preprint arXiv:2106.13797}, 2021.

\bibitem{wang2021pyramid}
Wenhai Wang, Enze Xie, Xiang Li, Deng-Ping Fan, Kaitao Song, Ding Liang, Tong
  Lu, Ping Luo, and Ling Shao.
\newblock Pyramid vision transformer: A versatile backbone for dense prediction
  without convolutions.
\newblock In {\em IEEE ICCV}, 2021.

\bibitem{wu2021cvt}
Haiping Wu, Bin Xiao, Noel Codella, Mengchen Liu, Xiyang Dai, Lu Yuan, and Lei
  Zhang.
\newblock Cvt: Introducing convolutions to vision transformers.
\newblock {\em arXiv preprint arXiv:2103.15808}, 2021.

\bibitem{xiao2018unified_upernet}
Tete Xiao, Yingcheng Liu, Bolei Zhou, Yuning Jiang, and Jian Sun.
\newblock Unified perceptual parsing for scene understanding.
\newblock In {\em Proceedings of the European Conference on Computer Vision
  (ECCV)}, pages 418--434, 2018.

\bibitem{yin2020disentangled_DNL}
Minghao Yin, Zhuliang Yao, Yue Cao, Xiu Li, Zheng Zhang, Stephen Lin, and Han
  Hu.
\newblock Disentangled non-local neural networks, 2020.

\bibitem{Yuan_2021_ICCV}
Li Yuan, Yunpeng Chen, Tao Wang, Weihao Yu, Yujun Shi, Zi-Hang Jiang,
  Francis~E.H. Tay, Jiashi Feng, and Shuicheng Yan.
\newblock Tokens-to-token vit: Training vision transformers from scratch on
  imagenet.
\newblock In {\em Proceedings of the IEEE/CVF International Conference on
  Computer Vision (ICCV)}, pages 558--567, October 2021.

\bibitem{zhang2021multi}
Pengchuan Zhang, Xiyang Dai, Jianwei Yang, Bin Xiao, Lu Yuan, Lei Zhang, and
  Jianfeng Gao.
\newblock Multi-scale vision longformer: A new vision transformer for
  high-resolution image encoding.
\newblock {\em ICCV 2021}, 2021.

\bibitem{zhang2021aggregating}
Zizhao Zhang, Han Zhang, Long Zhao, Ting Chen, and Tomas Pfister.
\newblock Aggregating nested transformers.
\newblock In {\em arXiv preprint arXiv:2105.12723}, 2021.

\bibitem{zhao2017pspnet}
Hengshuang Zhao, Jianping Shi, Xiaojuan Qi, Xiaogang Wang, and Jiaya Jia.
\newblock Pyramid scene parsing network.
\newblock In {\em CVPR}, 2017.

\bibitem{SETR}
Sixiao Zheng, Jiachen Lu, Hengshuang Zhao, Xiatian Zhu, Zekun Luo, Yabiao Wang,
  Yanwei Fu, Jianfeng Feng, Tao Xiang, Philip~H.S. Torr, and Li Zhang.
\newblock Rethinking semantic segmentation from a sequence-to-sequence
  perspective with transformers.
\newblock In {\em CVPR}, 2021.

\bibitem{ADE20K}
Bolei Zhou, Hang Zhao, Xavier Puig, Tete Xiao, Sanja Fidler, Adela Barriuso,
  and Antonio Torralba.
\newblock Semantic understanding of scenes through the ade20k dataset.
\newblock {\em International Journal of Computer Vision}, 127(3):302--321,
  2019.

\bibitem{zhou2021nnformer}
Hong-Yu Zhou, Jiansen Guo, Yinghao Zhang, Lequan Yu, Liansheng Wang, and Yizhou
  Yu.
\newblock nnformer: Interleaved transformer for volumetric segmentation, 2021.

\bibitem{zhu2020deformable}
Xizhou Zhu, Weijie Su, Lewei Lu, Bin Li, Xiaogang Wang, and Jifeng Dai.
\newblock Deformable detr: Deformable transformers for end-to-end object
  detection.
\newblock In {\em International Conference on Learning Representations}, 2020.

\end{thebibliography}
}

\end{document}